\documentclass{article}

\usepackage{arxiv}

\usepackage[utf8]{inputenc} 
\usepackage[T1]{fontenc}    
\usepackage{hyperref}       
\usepackage{url}            
\usepackage{booktabs}       
\usepackage{amsfonts}       
\usepackage{nicefrac}       
\usepackage{microtype}      
\usepackage{lipsum}
\usepackage{graphicx}
\usepackage{algorithmic}
\usepackage[tight,footnotesize]{subfigure}
\usepackage{caption}
\usepackage{adjustbox}
\usepackage{makecell} 
\usepackage[linesnumbered,ruled]{algorithm2e}

\usepackage{pifont}
\newcommand{\cmark}{\ding{51}}%
\newcommand{\xmark}{\ding{55}}%

\graphicspath{ {./images/} }

\title{Standardization of Neuromuscular Reflex Analysis — Role of Fine-Tuned Vision-Language Model Consortium and OpenAI gpt-oss Reasoning LLM Enabled Decision Support System}

\author{
Eranga Bandara \\
Old Dominion University \\
Norfolk, VA, USA \\
\texttt{cmedawer@odu.edu} \\
\And
Ross Gore \\
Old Dominion University \\
Norfolk, VA, USA \\
\texttt{rgore@odu.edu} \\
\And
Sachin Shetty \\
Old Dominion University \\
Norfolk, VA, USA \\
\texttt{sshetty@odu.edu} \\
\And
Ravi Mukkamala \\
Old Dominion University \\
Norfolk, VA, USA \\
\texttt{mukka@odu.edu} \\
\And
Christopher Rhea \\
Old Dominion University \\
Norfolk, VA, USA \\
\texttt{crhea@odu.edu} \\
\And
Amin Hass \\
AnaletIQ, \\
VA, USA \\
\texttt{aminhass@analetiq.com} \\
\And
Atmaram Yarlagadda \\
McDonald Army Health Center \\
Newport News, VA, USA \\
\texttt{atmaram.yarlagadda.civ@health.mil} \\
\And
Shaifali Kaushik \\
McDonald Army Health Center \\
Newport News, VA, USA \\
\texttt{Skaus2000@gmail.com} \\
\And
L.H.M.P.De Silva \\
University of Sri Jayewardenepura \\
Sri Lanka \\
\texttt{malith@sjp.ac.lk} \\
\And
Andriy Maznychenko \\
University of Gdańsk, \\
Poland \\
\texttt{andrii.maznychenko@awf.gda.pl} \\
\And
Inna Sokolowska \\
University of Gdańsk, \\
Poland \\
\texttt{inna.sokolowska@awf.gda.pl} \\
\And
Kasun De Zoysa \\
University of Colombo \\
Sri Lanka \\
\texttt{kasun@ucsc.cmb.ac.lk} \\
}

\begin{document}
\maketitle
\begin{abstract}
Accurate assessment of neuromuscular reflexes, such as the H-reflex, plays a critical role in sports science, rehabilitation, and clinical neurology. Traditional analysis of H-reflex EMG waveforms is subject to variability and interpretation bias among clinicians and researchers, limiting reliability and standardization. To address these challenges, we propose a Fine-Tuned Vision-Language Model (VLM) Consortium and a reasoning Large-Language Model (LLM)-enabled Decision Support System for automated H-reflex waveform interpretation and diagnosis. Our approach leverages multiple VLMs, each fine-tuned on curated datasets of H-reflex EMG waveform images annotated with clinical observations, recovery timelines, and athlete metadata. These models are capable of extracting key electrophysiological features and predicting neuromuscular states—including fatigue, injury, and recovery—directly from EMG images and contextual metadata. Diagnostic outputs from the VLM consortium are aggregated using a consensus-based method and refined by a specialized reasoning LLM, which ensures robust, transparent, and explainable decision support for clinicians and sports scientists. The end-to-end platform orchestrates seamless communication between the VLM ensemble and the reasoning LLM, integrating prompt engineering strategies and automated reasoning workflows using LLM Agents. Each VLM was fine-tuned using state-of-the-art techniques, including Low-Rank Adaptation (LoRA) and 4-bit quantization, enabling efficient deployment on consumer-grade hardware. Experimental results demonstrate that this hybrid system delivers highly accurate, consistent, and interpretable H-reflex assessments significantly advancing the automation and standardization of neuromuscular diagnostics. To our knowledge, this work represents the first integration of a fine-tuned VLM consortium with a reasoning LLM for image-based H-reflex analysis, laying the foundation for next-generation AI-assisted neuromuscular assessment and athlete monitoring platforms.
\end{abstract}


\section{Introduction}

Neuromuscular reflexes are essential elements of the human motor control system, enabling rapid, involuntary responses to sensory input that help maintain stability, coordination, and protection against injury~\cite{h-reflex-nuro}. Among these, the Hoffmann reflex (H-reflex) is a well-established electrophysiological measure for assessing the excitability and integrity of the spinal cord reflex arc. It is elicited through electrical stimulation of peripheral nerves and recorded via electromyography (EMG), providing a quantifiable indicator of neuromuscular pathway function~\cite{h-reflex-nuro}. Owing to its sensitivity and reproducibility, the H-reflex is extensively used in neurology, rehabilitation, sports science, and performance monitoring to evaluate recovery after injury, track neuromuscular disorders, and investigate adaptive changes in motor control~\cite{h-reflex-sport, h-reflex-f-wave}. Despite its importance, traditional approaches to H-reflex analysis rely primarily on visual inspection and manual quantification of EMG waveforms, or on semi-automated signal processing methods~\cite{related-work-4-emgllm}. These approaches, while effective in controlled environments, suffer from significant limitations~\cite{h-reflex-standardization}. Manual interpretation is susceptible to inter and intra-rater variability, potentially affecting the consistency and reliability of results. The time-intensive nature of manual analysis constrains throughput, making large-scale or real-time monitoring impractical. Furthermore, existing methods often fail to integrate the full spectrum of available metadata—such as patient history, training status, or contextual factors—thereby reducing the power for personalized diagnostics. Current semi-automated tools also tend to operate as black boxes, providing limited interpretability or flexibility in reasoning with complex, multi-modal datasets.

To address these challenges, we propose a novel platform for automated H-reflex analysis, built upon a Fine-Tuned Vision-Language Model Consortium and a reasoning language model-enabled Decision Support System~\cite{vision-language-model, vlm-image-classification}. Our approach leverages multiple VLMs, each fine-tuned on curated datasets of H-reflex EMG waveform images that are richly annotated with clinical observations, recovery timelines, and athlete metadata. These models are capable of extracting key electrophysiological features and accurately predicting neuromuscular states—including fatigue, injury, and recovery—directly from EMG images and contextual information. Diagnostic outputs generated by the VLM consortium are aggregated using a consensus-based method and subsequently refined by a specialized reasoning LLM~\cite{reasoning-llms, o3}, ensuring robust, transparent, and explainable decision support for clinicians, sports scientists, and researchers~\cite{h-reflex-sport}.

The end-to-end platform orchestrates seamless communication between the VLM ensemble and the reasoning LLM, integrating advanced prompt engineering strategies and automated reasoning workflows using LLM agents~\cite{agentic-ai, llm-agents}. Each VLM within the consortium is fine-tuned using Low-Rank Adapters (LoRA) and 4-bit quantization, enabling efficient training and deployment on consumer-grade hardware without sacrificing performance~\cite{lora, qlora}. Experimental results demonstrate that this hybrid system delivers highly accurate, consistent, and interpretable H-reflex assessments, significantly advancing both the automation and standardization of neuromuscular diagnostics. By automating labor-intensive analysis and integrating contextual metadata to enrich clinical and performance insights, our system supports informed decision-making across neurology, rehabilitation, and sports science~\cite{h-reflex-nuro}. To our knowledge, this work represents the first integration of a fine-tuned VLM consortium with a reasoning LLM for image-based H-reflex analysis, laying the foundation for next-generation AI-assisted neuromuscular assessment and athlete monitoring platforms. The following are our main contributions of this research.

\begin{enumerate}
\item Fine-tuning a consortium of Vision-Language Models to analyze H-reflex EMG waveform images and predict neuromuscular states such as fatigue, injury, and recovery.
\item Integrating a specialized reasoning LLM to refine and validate diagnostic outputs, ensuring robust, transparent, and explainable neuromuscular assessments based on the VLM consortium’s predictions.
\item Automating the end-to-end workflow for H-reflex analysis and neuromuscular diagnosis by orchestrating seamless communication between the VLM ensemble and the reasoning LLM, facilitated by AI agents and advanced prompt engineering.
\item Implementing and validating a prototype of the NeuroLens platform, integrating multiple fine-tuned VLMs with the reasoning LLM, and demonstrating its effectiveness for standardized and scalable neuromuscular reflex assessment in clinical and sports science contexts.
\end{enumerate}

The remainder of the paper is organized as follows: Section 2 introduces the core technologies that underpin the proposed AI-assisted neuromuscular reflex analysis platform. Section 3 details the overall system architecture, highlighting the integration of VLMs and reasoning engines for H-reflex waveform interpretation. Section 4 outlines the platform's core functionalities and operational workflow, from data ingestion to final reflex response interpretation. Section 5 presents implementation details and evaluates the system’s performance across neuromuscular analysis tasks. Section 6 reviews related work and contextualizes our approach within the broader landscape of AI-driven sports science and neuromuscular assessment systems. Finally, Section 7 concludes the paper and discusses potential directions for future research, performance monitoring applications, and integration into athlete recovery management workflows.

\begin{figure}[h]
\centering{}
\includegraphics[width=5.2in]{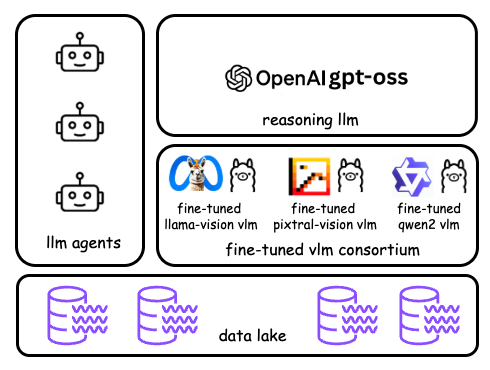}
\vspace{-0.1in}
\DeclareGraphicsExtensions.
\caption{Platform architecture.}
\label{indy528-architecture}
\end{figure}

\begin{figure}[h]
\centering{}
\includegraphics[width=5.2in]{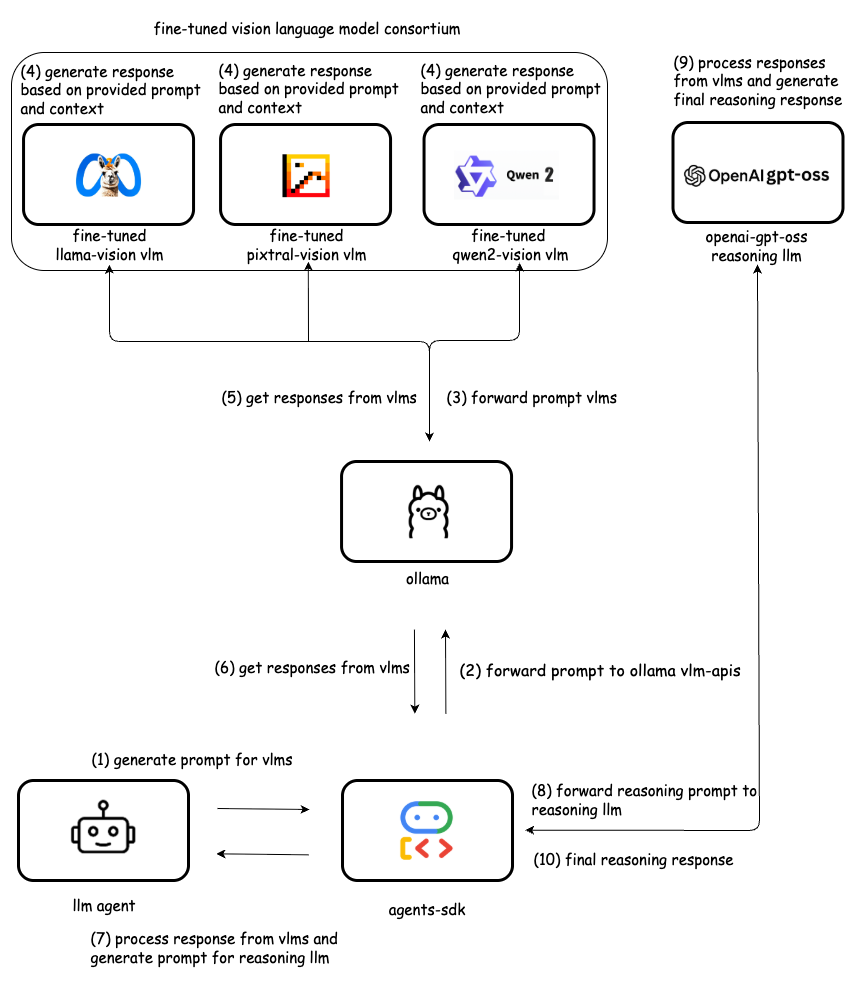}
\vspace{-0.1in}
\DeclareGraphicsExtensions.
\caption{LLM integration flow with Ollama LLM-API}
\label{llama2-flow}
\end{figure}

\section{Background}

This section provides a foundational overview of the core scientific and technological concepts underpinning the proposed AI-assisted neuromuscular reflex analysis platform. In particular, we highlight the basis and clinical importance of the H-reflex in neuromuscular analysis, recent advancements in Large Language Models (LLMs), reasoning-capable LLMs, fine-tuning techniques, and the emerging paradigm of AI agents.

\subsection{The H-reflex and Neuromuscular Diagnostics}

The Hoffmann reflex (H-reflex) is a fundamental neurophysiological marker used to assess the excitability and integrity of the monosynaptic reflex arc within the human neuromuscular system~\cite{h-reflex-nuro}. By electrically stimulating a peripheral nerve and recording the resultant electromyographic (EMG) responses in a target muscle, the H-reflex enables noninvasive quantification of spinal cord and motor neuron function. Typically, a single supramaximal stimulus produces both a direct motor response (M-wave) and the H-reflex, which is mediated via Ia afferent fibers synapsing on alpha motor neurons in the spinal cord.

The amplitude, latency, and recruitment properties of the H-reflex waveform provide sensitive indicators of neuromuscular health, making it a valuable tool in clinical neurophysiology, rehabilitation, and sports science~\cite{h-reflex-sport}. In athletes, longitudinal monitoring of the H-reflex supports objective evaluation of fatigue, recovery status, training adaptations, and neuromuscular injury risk~\cite{h-reflex-f-wave}. It is also widely employed to investigate pathologies affecting motor neuron excitability, such as neuropathies, spinal cord injuries, and neurodegenerative disorders. However, traditional H-reflex analysis relies on manual or semi-automated interpretation of EMG waveforms, a process that is labor-intensive, time-consuming, and subject to inter-rater variability~\cite{realted-work-1-emg-sensor}. Additionally, the integration of contextual metadata—such as athlete characteristics, clinical observations, and recovery timelines—is rarely standardized, limiting the generalizability and clinical utility of reflex-based assessments.

\subsection{Vision-Language Models (VLMs)}

Vision-Language Models are advanced deep neural networks trained on both large-scale text and image datasets, enabling them to jointly interpret, generate, and reason across visual and textual modalities. These models form the foundation of modern multi-modal AI systems~\cite{vistion-language-model-comparison}, and have demonstrated exceptional performance in tasks such as image captioning, visual question answering, medical image interpretation, and cross-modal retrieval~\cite{vlm-image-classification}.

Several prominent VLMs—such as Llama-Vision~\cite{llama-3, llama-4}, Pixtral-Vision~\cite{mistral-fine-tune}, and Qwen2-VL~\cite{qwen2} available alongside language-focused proprietary models like OpenAI’s GPT~\cite{gpt-llm}, Google’s Gemini~\cite{gemini}. Open-source VLMs offer substantial benefits for healthcare and biomedical applications, including transparency, customizability, and cost-effective deployment. For instance, Llama-Vision~\cite{llama-vision, proof-of-tbi} and Pixtral-Vision~\cite{mistral-fine-tune} provide strong performance with efficient architectures suitable for integration with visual processing modules. Many modern VLMs are optimized for multi-lingual, on-device, or edge deployment, supporting scalable and privacy-preserving applications in clinical and research environments.

\subsection{Reasoning LLMs}

While foundational LLMs excel in pattern recognition and natural language generation, they often lack the capacity for structured, multi-step reasoning. Reasoning LLMs\cite{reasoning-llms} address this limitation by being specifically designed or fine-tuned to synthesize diverse inputs, resolve conflicting information, and support logical decision-making processes. Unlike traditional LLMs that primarily rely on next-token prediction, reasoning models simulate higher-order cognitive functions akin to human deductive reasoning\cite{o3}.

OpenAI-gpt-oss~\cite{gpt-oss} is an open-source reasoning LLM designed to perform advanced evaluative and comparative tasks across multiple inputs. Unlike traditional generative LLMs that focus on single-output prediction, OpenAI-gpt-oss is capable of synthesizing responses, resolving contradictions, and applying logical inference to arrive at consistent, well-reasoned conclusions. It excels in tasks involving multi-model output reconciliation, ranking, and consensus generation. gpt-oss also supports chain-of-thought reasoning, tool invocation, and visible reasoning steps for improved transparency and auditability~\cite{cot-reasoning}. These properties make it ideally suited for structured, multi-step interpretative tasks such as aggregating and reasoning over outputs from vision-language models in neuromuscular reflex analysis.


\subsection{VLM Fine-tuning}

Fine-tuning is a key technique for adapting pre-trained VLMs to specialized downstream tasks and domains. It involves retraining the model on curated, task-specific datasets that combine both visual (e.g., EMG waveform images) and textual (e.g., athlete metadata, clinical observations) inputs~\cite{bassa-llama, vindsec-llama}. This process allows the VLM to learn domain-relevant associations and produce outputs that are precisely aligned with neuromuscular reflex analysis and related biomedical applications~\cite{mistral-fine-tune}.

To optimize the efficiency and scalability of fine-tuning, Low-Rank Adapters (LoRA)\cite{lora} are commonly employed. LoRA introduces trainable low-rank matrices into the transformer architecture, allowing for efficient, task-specific adaptation while significantly reducing the number of trainable parameters. For resource-constrained settings, Quantized LoRA (QLoRA)\cite{qlora} provides even greater memory and compute efficiency by quantizing model weights to 4-bit representations, while retaining nearly full-precision performance. These techniques collectively enable the practical fine-tuning of large VLMs on modest hardware, making advanced multi-modal models accessible for clinical and research applications.

Several open-source libraries facilitate efficient fine-tuning workflows for VLMs. For example, Unsloth~\cite{llamafactory-unsloth} provides high-speed, memory-efficient fine-tuning for models such as Llama-Vision~\cite{llama-3, llama-4}, Pixtral-Vision~\cite{mistral-fine-tune}, and Qwen2~\cite{qwen2}, leveraging LoRA and QLoRA methods. It supports both consumer-grade GPUs (e.g., NVIDIA RTX 3090) and scalable cloud environments, including TPU-enabled platforms like Google Colab~\cite{google-tpu}. Successful fine-tuning of VLMs typically requires GPUs with ample VRAM and compute capabilities. High-performance GPUs such as the NVIDIA A100 and H100 are ideal for large-scale training, while more accessible hardware like the NVIDIA RTX 3090/4090 and Tesla T4 are suitable for small to medium-scale fine-tuning and rapid prototyping~\cite{a100-gpu}.

\subsection{AI Agents and Agentic AI}

AI agents are autonomous computational entities designed to perform complex tasks by interacting with data sources, machine learning models, and external APIs within dynamic or uncertain environments. When these agents are powered by LLMs, they are referred to as LLM agents, capable of interpreting natural language instructions, generating structured outputs, managing tasks, and coordinating actions across digital ecosystems~\cite{llm-agents, agentic-ai}.

Agentic AI extends this concept by organizing multiple LLM agents into collaborative, role-specialized systems that demonstrate advanced capabilities such as long-term planning, self-reflection, adaptive behavior, and multi-agent coordination~\cite{agentic-ai}. These systems operate through agent hierarchies or workflows in which each agent performs a specific role, such as prompt engineering, retrieval, inference, evaluation, or integration. The modularity of agentic architectures enhances scalability, interpretability, and reusability, making them particularly suitable for domains requiring structured reasoning, task delegation, and reliable decision support.

\section{System Architecture}

Figure~\ref{indy528-architecture} describes the architecture of the platform. The proposed platform is composed of 4 layers: 1) Data Lake layer, 2) LLM Agent Layer, 3) VLM Layer, and 4) Reasoning Layer. Below is a brief description of each layer.

\subsection{Data Lake Layer}

The Data Lake layer serves as the foundational infrastructure for managing and storing the diverse, large-scale datasets essential for automated neuromuscular reflex analysis. This centralized repository is designed to support the training and fine-tuning of VLMs and reasoning language models by aggregating a wide array of multimodal data relevant to H-reflex diagnostics~\cite{h-reflex-f-wave, llama-recipe}. The Data Lake hosts collections of annotated EMG waveform images, corresponding athlete metadata (such as age, gender, sport, and training context), clinical observations, recovery timelines, and injury histories. These richly labeled datasets enable the platform to capture the complex physiological, contextual, and temporal variability inherent in neuromuscular assessments~\cite{h-reflex-nuro}. By centralizing and standardizing this information, the Data Lake layer empowers the development of robust, generalizable AI models capable of accurate, explainable, and individualized interpretation of neuromuscular reflex data across diverse populations and use cases, from clinical rehabilitation to elite sports performance monitoring.

\subsection{LLM Agent Layer}

The LLM Agent Layer serves as the orchestration and automation core of the platform, enabling seamless integration and coordination across the Data Lake, fine-tuned VLMs, and the OpenAI-o3 reasoning engine. In this layer, LLM agents act as orchestrators responsible for custom prompt engineering, ensuring efficient communication between all components and supporting the end-to-end automation of neuromuscular reflex analysis. Specifically, the LLM agents dynamically construct prompts using EMG waveform images and associated metadata—such as athlete characteristics, clinical observations, and recovery timelines—retrieved from the Data Lake~\cite{prompt-engineering}. These prompts are used to query the ensemble of fine-tuned VLMs, each of which outputs preliminary assessments regarding neuromuscular state, fatigue, injury, or recovery based on both the visual and contextual information.

The agents then aggregate these VLM outputs and format them into consolidated, structured prompts tailored for the OpenAI-o3 reasoning LLM~\cite{o3}. Leveraging its advanced reasoning capabilities, the OpenAI-o3 model evaluates and synthesizes the collective outputs of the VLM consortium to generate a refined, explainable, and clinically relevant interpretation of the H-reflex data. By adapting prompts to match the input requirements and context of each model, the LLM Agent Layer ensures optimal information flow, interoperability, and consistency throughout the workflow. This orchestrated process not only enhances the accuracy and transparency of neuromuscular assessments but also enables a fully automated, end-to-end AI-driven diagnostic system, as illustrated in Figure~\ref{llama2-flow}.

\subsection{VLM Layer}

The VLM Layer serves as the analytical core of the platform, enabling the system to interpret complex neuromuscular signals and generate accurate, explainable assessments. This layer comprises a consortium of fine-tuned VLMs, each trained on domain-specific datasets of annotated H-reflex EMG waveform images, athlete metadata, and clinical observations~\cite{llm-finetune, llama-recipe}. These models are specialized to extract and analyze key electrophysiological features, as well as contextual information, to assess neuromuscular states such as fatigue, injury, and recovery. The fine-tuned VLMs are deployed and managed using efficient frameworks optimized for scalable inference and deployment on consumer-grade hardware, ensuring the platform can maintain high performance and accessibility across diverse settings.

As illustrated in Figure~\ref{llama2-flow}, the LLM Agent Layer interfaces with the VLM consortium, orchestrating prompt generation, model invocation, and aggregation of preliminary assessments. By leveraging multiple specialized models within the consortium, the VLM Layer enhances the robustness and reliability of neuromuscular analysis through diversity in visual reasoning and interpretation. This collaborative approach supports a more comprehensive and consistent assessment of H-reflex waveforms and related clinical outcomes, advancing the automation and standardization of neuromuscular diagnostics in both clinical and sports science domains.


\begin{figure}[h]
\centering{}
\includegraphics[width=5.2in]{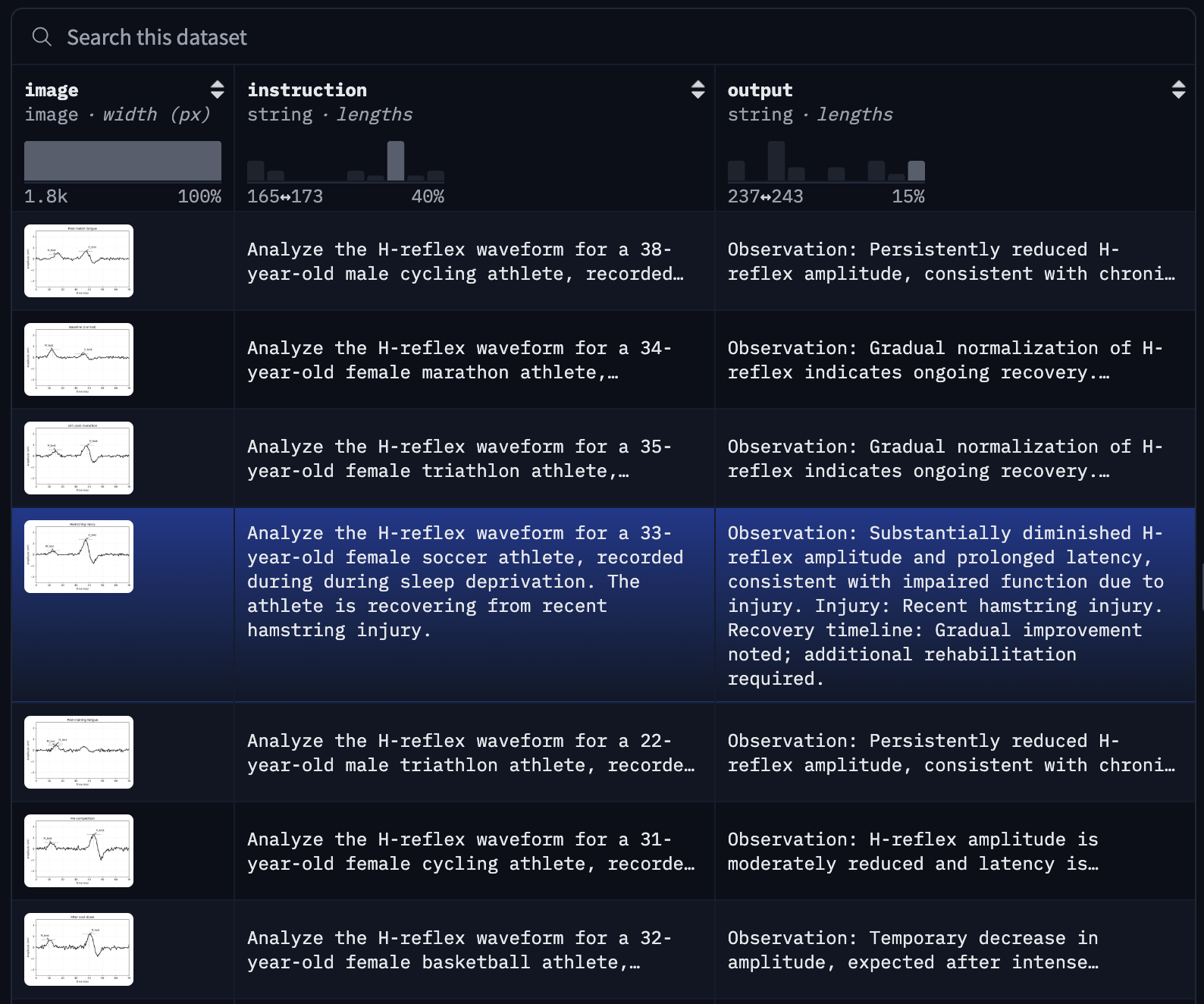}
\DeclareGraphicsExtensions.
\caption{Representative samples from the H-reflex neuromuscular dataset used to fine-tune the VLMs.}
\label{dataset-format}
\end{figure}

\begin{figure}[h]
\centering{}
\includegraphics[width=5.2in]{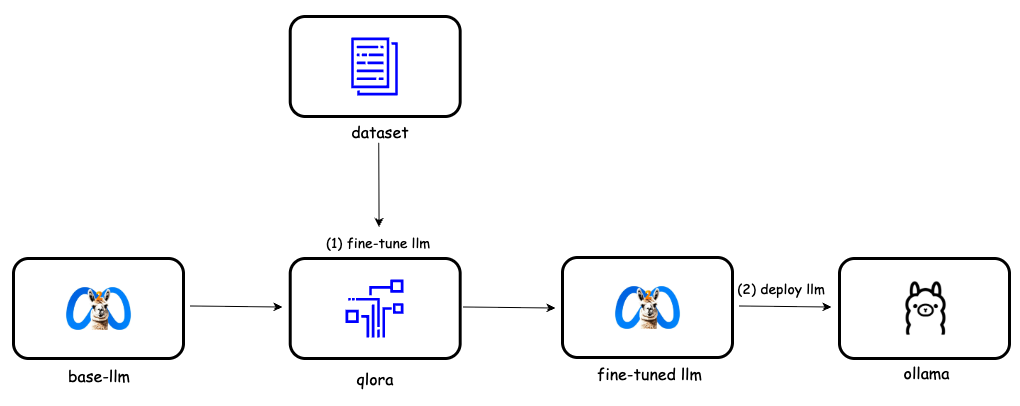}
\vspace{-0.1in}
\DeclareGraphicsExtensions.
\caption{Fine-tune VLM with Qlora and deploy with Ollama.}
\label{llm-fine-tune}
\end{figure}

\subsection{Reasoning LLM Layer}

The Reasoning Layer embodies the platform’s advanced cognitive and decision-making capabilities, leveraging state-of-the-art reasoning language models to synthesize nuanced clinical insights. The Reasoning LLM acts as the cognitive and synthesis engine of the platform, responsible for high-level reasoning, integration, and refinement of neuromuscular assessment predictions generated by the VLM consortium.

Within the platform, OpenAI-gpt-oss~\cite{gpt-oss} is used as the reasoning LLM and serves as the final decision-making engine. It receives preliminary neuromuscular assessments and diagnostic predictions from the ensemble of fine-tuned VLMs, then performs structured reasoning to evaluate, cross-validate, and refine these outputs~\cite{llm-reasoning, proof-of-tbi}. By synthesizing diverse model perspectives, each based on different EMG waveform features, athlete metadata, and contextual information, the reasoning LLM determines the most consistent and clinically relevant interpretation of the H-reflex data, supporting accurate and explainable outcomes for fatigue, injury, and recovery status. The LLM Agent Layer facilitates this process by aggregating and formatting the VLM outputs into structured, context-aware prompts tailored for the reasoning LLM to process heterogeneous inputs and deliver a final, consensus-driven assessment.

By integrating probabilistic reasoning, consistency checks, and domain knowledge, the Reasoning LLM Layer plays a pivotal role in enhancing the reliability, transparency, and clinical utility of AI-assisted neuromuscular reflex analysis across both clinical and sports science applications.

\section{Platform Functionality}

There are four main functionalities of the platform: 1) Data Lake Setup, 2) VLM Fine-Tuning, 3) Prediction of Fine-tuned VLMs, and 4) Final Prediction by Reasoning LLM. This section goes into the specifics of these functions.

\subsection{Data Lake Setup}

The first step in the platform’s workflow involves the setup of the Data Lake, which serves as the foundational layer for storing, managing, and accessing large-scale multimodal datasets essential for neuromuscular reflex analysis. These datasets include annotated EMG waveform images, athlete and session metadata (such as age, gender, sport, and training context), clinical observations, recovery timelines, and records of injuries or interventions. This comprehensive and centralized repository supports the training and fine-tuning of VLMs and reasoning models that underpin the platform’s predictive capabilities~\cite{vlm-image-classification}.

All data stored in the Data Lake is standardized and richly labeled, enabling the platform to capture the complex variability inherent in neuromuscular assessments across different populations and scenarios. By providing a robust, scalable, and secure data infrastructure, the Data Lake facilitates the development of fine-tuned models capable of interpreting subtle changes in H-reflex signals, understanding contextual factors, and supporting consistent, data-driven neuromuscular assessments. This infrastructure is a critical enabler for scalable, explainable, and automated interpretation of neuromuscular reflex data in both clinical and sports performance contexts~\cite{h-reflex-sport}.

\subsection{VLM Fine-Tuning}

The second step in the platform workflow involves fine-tuning VLMs using the curated and pre-processed data stored in the Data Lake. This stage is crucial for transforming general-purpose models into specialized agents capable of interpreting H-reflex EMG waveform images, integrating athlete metadata, and generating context-aware neuromuscular predictions. Multiple state-of-the-art models—including Llama-Vision~\cite{llama-3, llama-4, wedagpt}, Pixtral-Vision~\cite{mistral-fine-tune}, and Qwen2~\cite{qwen2}—are fine-tuned on this domain-specific, multimodal dataset to adapt them to the complex physiological and contextual characteristics of neuromuscular assessments. The structure and composition of the dataset used for fine-tuning are illustrated in Figure~\ref{dataset-format}.

The fine-tuning process is carried out using the Unsloth library~\cite{llamafactory-unsloth}, which enables efficient large-scale adaptation of LLMs and VLMs. To ensure models are deployable on consumer-grade hardware without compromising performance, the process incorporates Quantized Low-Rank Adapters (QLoRA)\cite{qlora} with 4-bit quantization, as depicted in Figure\ref{llm-fine-tune}. This optimization significantly reduces memory and computational requirements, supporting real-time inference and deployment at scale.

Upon completion, the fine-tuned and quantized models are deployed using lightweight frameworks optimized for efficient inference, such as Ollama~\cite{ollama}. These specialized models form the analytical core of the platform, each capable of analyzing EMG waveform images and associated metadata to produce preliminary neuromuscular state predictions—such as fatigue level, injury status, or recovery progression—based on learned physiological patterns and contextual cues.

\subsection{Prediction by Fine-tuned VLMs}

Following the fine-tuning process, the next phase of the platform involves generating preliminary neuromuscular assessments and predictions using the consortium of fine-tuned VLMs. When new EMG waveform images and associated metadata are ingested, the platform’s LLM Agent Layer initiates the predictive analysis by interfacing with the specialized models through efficient inference frameworks such as Ollama~\cite{ollama}. To facilitate accurate and context-aware predictions, the LLM Agent employs custom prompt engineering, embedding the relevant waveform data, athlete characteristics, and contextual information into tailored prompts for each model~\cite{prompt-engineering-rag}. These prompts are carefully designed to match the input requirements of each VLM and to provide a comprehensive representation of the physiological and situational context.

Each fine-tuned model then analyzes the input, extracts key electrophysiological features and contextual signals, and produces its own prediction regarding neuromuscular state—such as fatigue level, injury risk, or stage of recovery. The individual outputs are collected by the LLM Agent, which organizes them into a structured format for downstream reasoning. This step ensures that the diverse analytical capabilities of the fine-tuned models are fully leveraged, providing rich, reliable, and interpretable insights into the current neuromuscular condition.

By enabling multiple independent evaluations across the model consortium, this layer enhances the diversity, robustness, and generalizability of the platform’s predictions, supporting consistent and nuanced assessments across a wide range of athletes and scenarios.

\subsection{Final Prediction by OpenAI-o3 Reasoning LLM}

To ensure the highest level of prediction accuracy, reliability, and contextual validity, the platform employs a consensus-based decision-making mechanism for final neuromuscular assessment. Rather than relying on the output of a single model, the platform aggregates predictive outputs from multiple fine-tuned VLMs within the consortium. These individual results are then evaluated, compared, and synthesized by  OpenAI-gpt-oss, a specialized reasoning LLM designed to perform advanced analytical inference~\cite{llm-reasoning}. As a core component of the architecture, OpenAI-gpt-oss serves as an intelligent adjudicator, capable of contextualizing, validating, and refining the predictions provided by the underlying VLMs. Leveraging its advanced reasoning capabilities, OpenAI-gpt-oss identifies the most consistent and contextually appropriate outcome from the diverse set of model-generated insights.

\begin{figure}[h]
\centering{}
\includegraphics[width=5.2in]{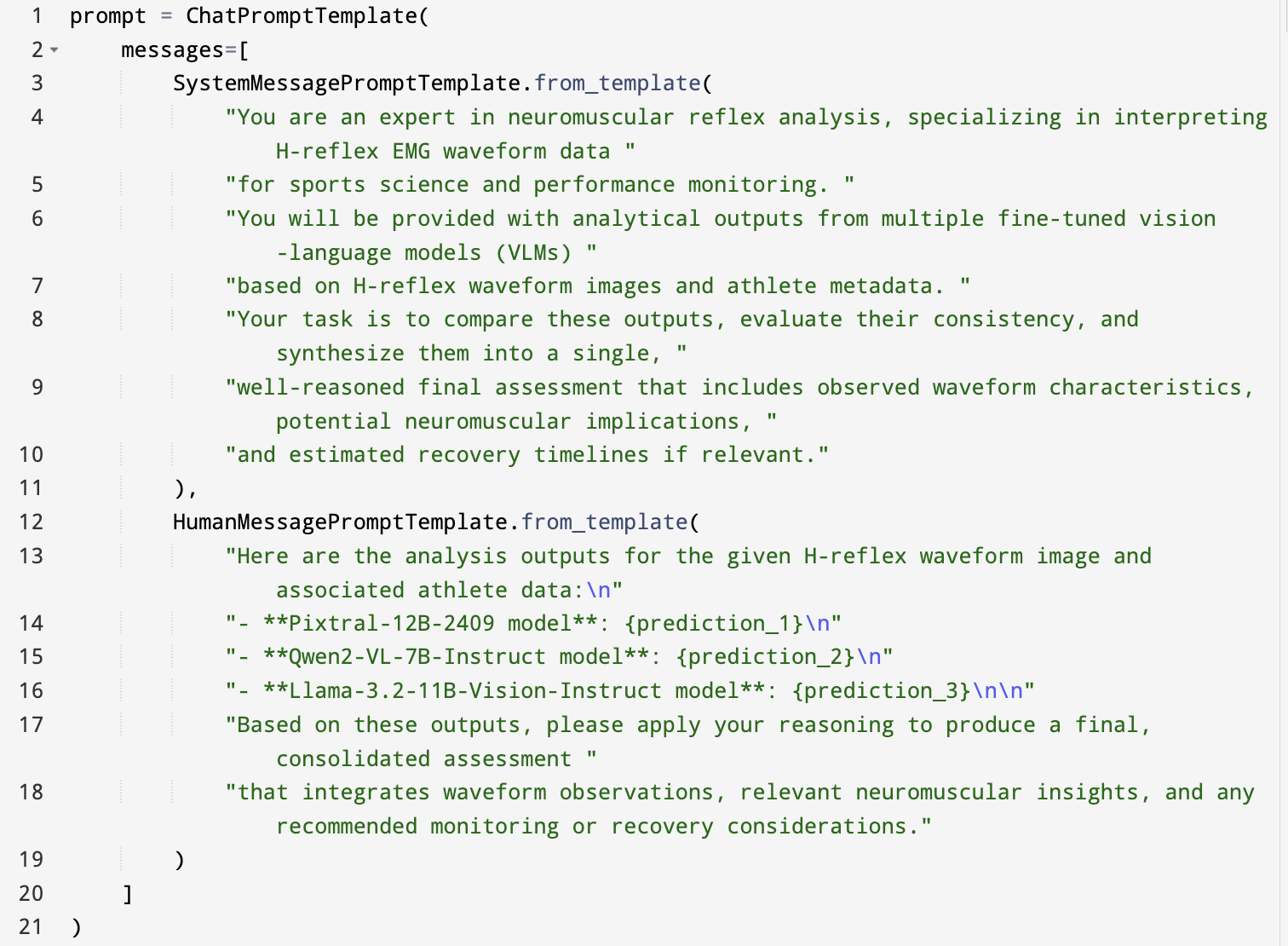}
\DeclareGraphicsExtensions.
\caption{Prompt for OpenAI-gpt-oss reasoning LLM for final prediction reasoning.}
\label{prompt}
\end{figure}

To enable this reasoning process, the LLM Agent constructs custom, structured prompts by embedding and organizing the outputs from the fine-tuned models. These prompts, as illustrated in Figure~\ref{prompt}, provide OpenAI-gpt-oss with a unified view of candidate assessments, associated waveform features, athlete metadata, and contextual cues. The reasoning LLM processes this composite input and produces a final consensus assessment that reflects a well-supported interpretation of the H-reflex data in relation to neuromuscular state, fatigue, injury risk, and recovery progression.

This consensus-driven architecture significantly improves the robustness and generalizability of predictive outputs by mitigating the limitations of individual models and reducing variability. By orchestrating this process through a transparent, explainable pipeline, the platform not only increases trustworthiness but also establishes a replicable framework for AI-assisted neuromuscular reflex interpretation and sports performance monitoring.

The integration of ensemble-based inference with symbolic reasoning marks a transformative shift in neuromuscular analytics, offering a scalable and interpretable decision support tool for clinicians, trainers, and researchers. It demonstrates the potential of combining large-scale vision-language understanding with structured reasoning to improve prediction and monitoring in complex physiological domains.

\begin{figure}[h]
\centering{}
\includegraphics[width=5.2in]{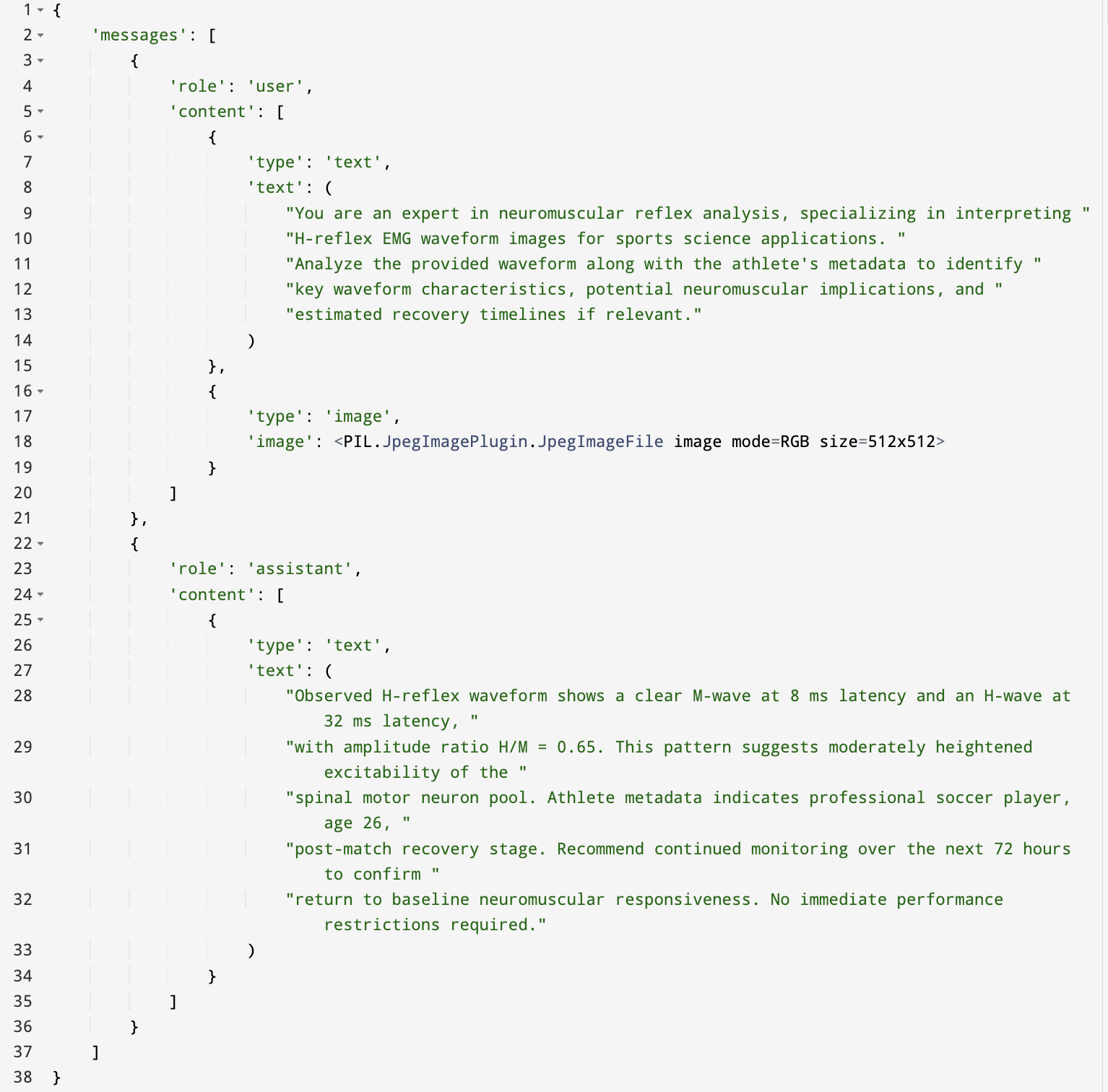}
\DeclareGraphicsExtensions.
\caption{The required data format of the unsloth library to fine-tune the VLMs.}
\label{unsloth-format}
\end{figure}

\begin{figure}[h]
\centering{}
\includegraphics[width=5.0in]{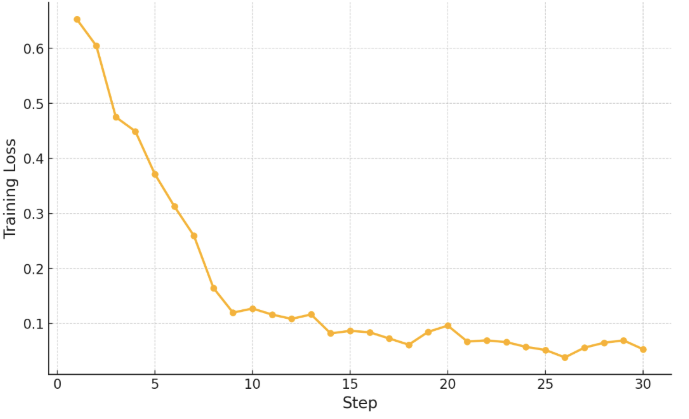}
\DeclareGraphicsExtensions.
\caption{Training loss during fine-tuning of the Llama-3.2-11B-Vision-Instruct LLM}
\label{unsloth-tranning-loss}
\end{figure}

\begin{figure}[h]
\centering{}
\includegraphics[width=5.0in]{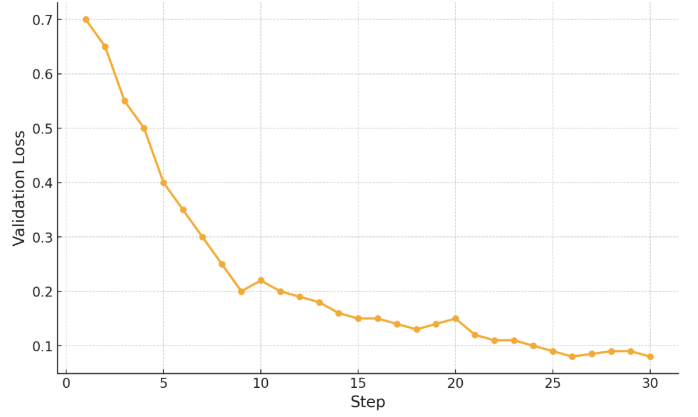}
\DeclareGraphicsExtensions.
\caption{Validation loss during fine-tuning of the Llama-3.2-11B-Vision-Instruct LLM}
\label{unsloth-validation-loss}
\end{figure}

\begin{figure}[h]
\centering{}
\includegraphics[width=5.0in]{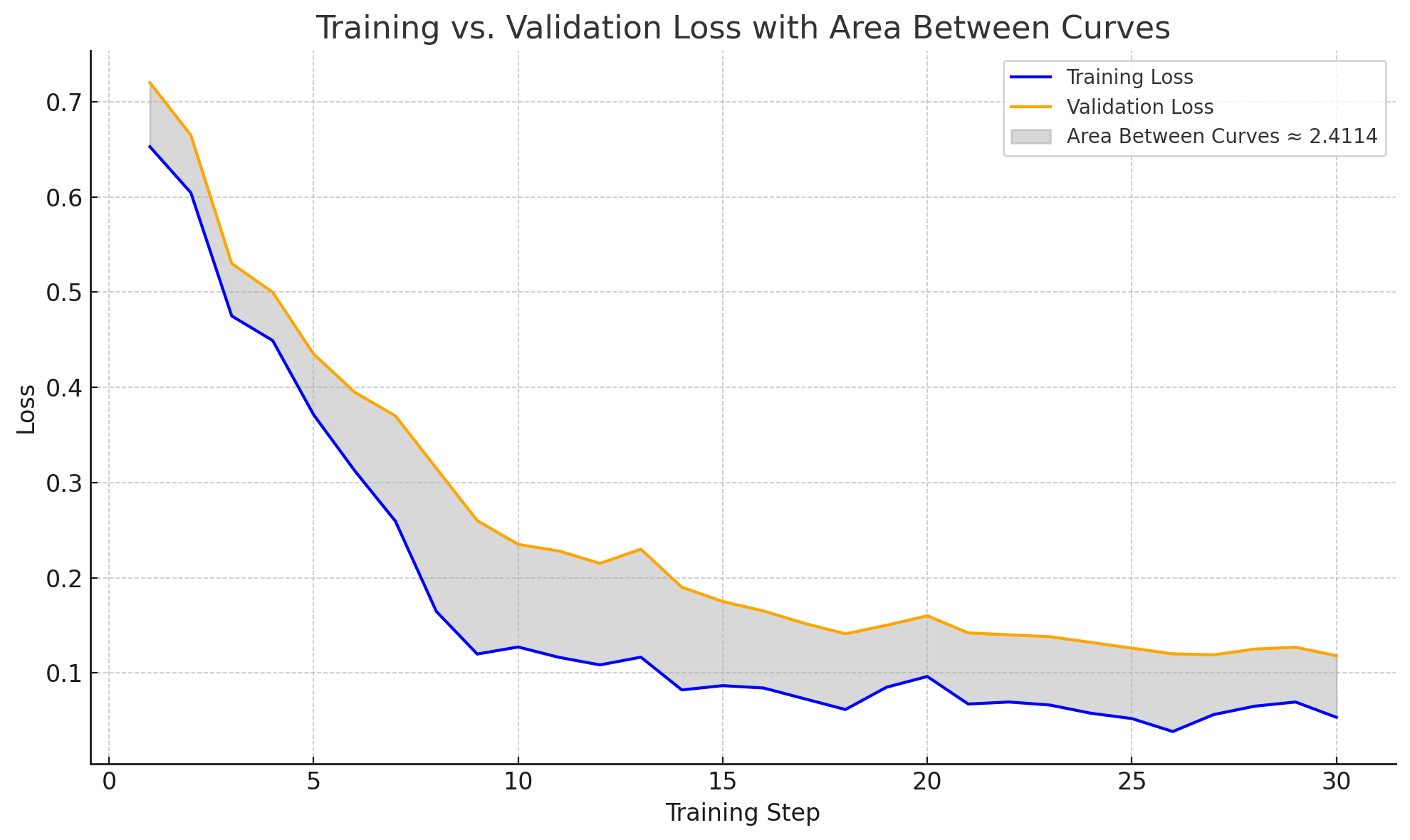}
\DeclareGraphicsExtensions.
\caption{Training vs. Validation Loss and Area Between Curves during Fine-Tuning of the Llama-3.2-11B-Vision-Instruct LLM}
\label{unsloth-tranning-validation-loss}
\end{figure}

\begin{figure}[h]
\centering{}
\includegraphics[width=5.0in]{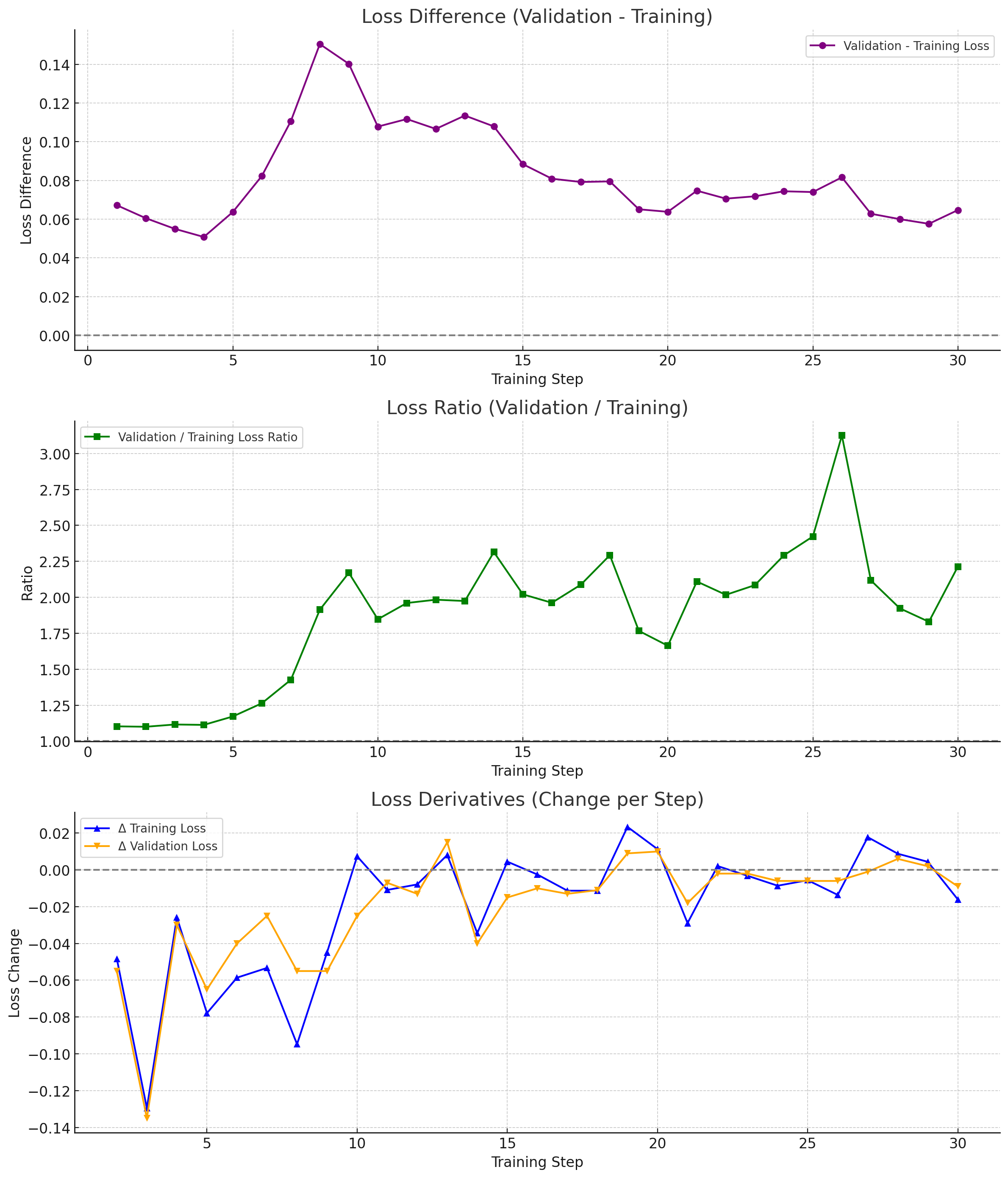}
\DeclareGraphicsExtensions.
\caption{Ratio of training to validation loss during the fine-tuning of the Llama-3.2-11B-Vision-Instruct LLM.}
\label{unsloth-loss-ratio}
\end{figure}

\section{Implementation and Evaluation}

The implementation of the proposed platform was carried out using three fine-tuned VLMs—Llama-Vision, Mistral-Vision, and Qwen2-VL~\cite{llama-3, vistion-language-model-comparison, on-device-qwen2}—in combination with the OpenAI-gpt-oss~\cite{reasoning-llms, o3, gpt-oss} reasoning LLM. The LLM Agent Layer was implemented using the OpenAI Agents SDK~\cite{openai-agent-sdk} and Google Agent Development Kit~\cite{agent-survey}, enabling secure orchestration, transparent auditability, and decentralized control of all model interactions.

Fine-tuning was conducted using the Unsloth library~\cite{llamafactory-unsloth} on Google Colab, leveraging both NVIDIA A100 GPUs and Tesla TPUs~\cite{google-tpu} to support efficient and scalable training cycles. The dataset consisted of approximately 1,200 annotated records, each containing an H-reflex EMG waveform image, comprehensive athlete metadata, contextual details (e.g., training or recovery phase), and expert-annotated observations or predictions regarding neuromuscular state, fatigue, injury risk, or recovery progression. These records were compiled from multiple data sources, as illustrated in Figure~\ref{dataset-format}.

The Unsloth framework requires input data to be structured in an instruction-based format~\cite{llamafactory-unsloth}. To meet this requirement, the dataset was preprocessed and transformed into the required schema, shown in Figure~\ref{unsloth-format}. Each training sample included fields such as instruction (providing the analysis context and metadata), image (representing the EMG waveform), and output (containing the model’s expected neuromuscular assessment or prediction). The dataset was partitioned into training, validation, and testing subsets using a 2/3, 1/6, 1/6 split, respectively. The training process was completed in approximately 1,627 seconds (27.12 minutes). Peak memory reservation during training was 14.605 GB, with actual memory utilization reaching 5.853 GB, equivalent to 39.69\% of reserved memory and 99.03\% of peak allocation. These results demonstrate that fine-tuning VLMs for neuromuscular reflex assessment using structured, multimodal data can be performed efficiently, even on moderate-scale datasets and accessible hardware. This underscores the practicality and accessibility of applying advanced AI methods in specialized biomedical domains.

After fine-tuning, the models were quantized using QLoRA~\cite{qlora}, enabling efficient operation on consumer-grade hardware. This optimization was essential for deploying the fine-tuned models with Ollama, a framework designed for lightweight yet high-performance model execution. Based on the predictions of the VLMs, the OpenAI-gpt-oss LLM synthesizes the collective outputs to generate a final consensus assessment. Custom prompts are used to instruct the OpenAI-gpt-oss Reasoning LLM, providing the necessary context for effective integration and reasoning across model outputs.

Platform performance was evaluated in three main areas: 1) Efficiency and accuracy of VLM fine-tuning, 2) Predictive performance and consistency of the VLM consortium, and 3) Advanced reasoning and consensus-building capabilities of the OpenAI-gpt-oss LLM. Results indicate that the proposed architecture enables robust, transparent, and scalable neuromuscular reflex assessment, setting the stage for widespread adoption in clinical and sports science applications.

\subsection{Evaluation of VLM Fine-Tuning}

This evaluation focuses on measuring the effectiveness of the fine-tuning process in improving the performance of VLMs in analyzing neuromuscular reflex imagery, specifically H-reflex waveforms. We evaluated the fine-tuned Llama-Vision model’s ability to interpret visual H-reflex data and produce structured observations on reflex amplitude, latency, and recovery status based on the provided image inputs~\cite{training-validation-loss}.

Throughout the fine-tuning process, we continuously monitored critical training metrics—specifically, training loss and validation loss—to assess the model's learning dynamics and generalization ability~\cite{llm-finetune}. As shown in Figure~\ref{unsloth-tranning-validation-loss}, the training loss (Figure~\ref{unsloth-tranning-loss}) and validation loss (Figure~\ref{unsloth-validation-loss}) both exhibit a steep decline during the initial training steps, indicating rapid adaptation to neuromuscular domain-specific patterns. The validation loss continues to decrease smoothly over time, stabilizing around step 25, which suggests improved generalization to unseen H-reflex samples. Meanwhile, the training loss decreases more aggressively and plateaus slightly earlier, signaling convergence. Figure~\ref{unsloth-tranning-validation-loss} also provides an integrated view of both metrics along with the area between the curves, representing the generalization gap. The relatively narrow and consistently shrinking gap further confirms the model's ability to generalize well without overfitting.

Figure~\ref{unsloth-loss-ratio} captures multiple key training dynamics, including the loss difference, loss ratio, and loss derivatives over training steps, offering valuable insights into the model’s convergence behavior. The consistently positive loss difference (validation loss exceeding training loss) suggests minor signs of overfitting at certain points, especially at steps with noticeable spikes. The loss ratio, ranging from 1.0 to 3.0, reflects varying degrees of generalization, where lower ratios indicate better alignment between training and validation performance. Additionally, the loss derivatives reveal rapid initial improvements followed by smaller, oscillating changes, indicating stabilization and saturation in the model’s learning process~\cite{llm-loss-ratio}.

These trends collectively indicate that the fine-tuning process was both effective and stable, enabling the VLM to adapt precisely to the neuromuscular reflex analysis domain while maintaining strong performance on unseen reflex waveform data.

\subsection{Prediction Performance of Fine-Tuned VLM Consortium}

Following the training phase, we evaluated the predictive performance of the fine-tuned VLMs in the context of neuromuscular reflex assessment. This evaluation compared expert-validated H-reflex observations—derived from electrophysiological waveform analysis—with the predictions generated by both the baseline (pre-trained) VLMs and their fine-tuned counterparts. The assessment focused on each model’s ability to accurately describe waveform characteristics, infer potential neuromuscular implications, and provide contextually appropriate recovery timelines.

\begin{figure}[h]
\centering{}
\includegraphics[width=5.2in]{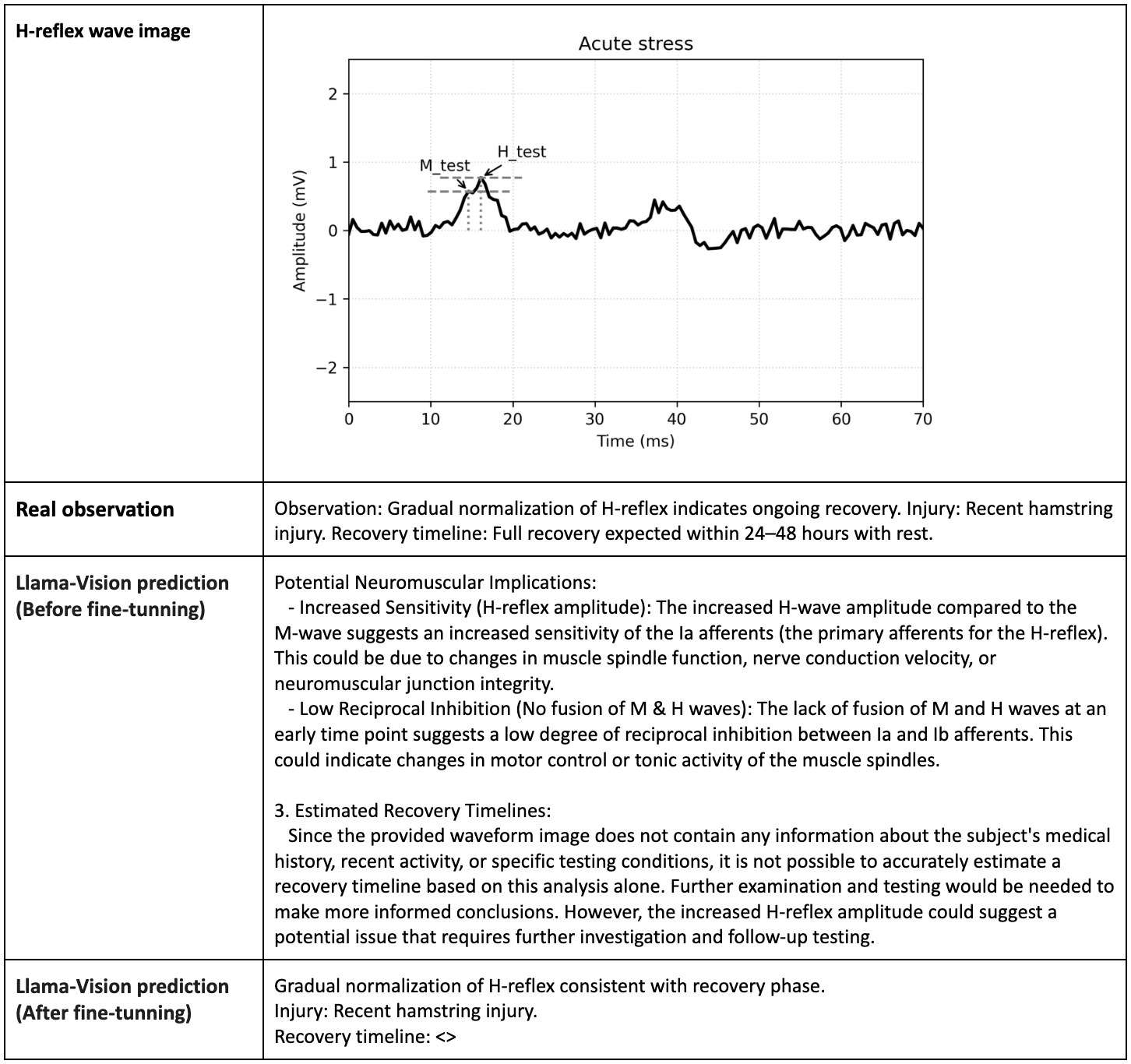}
\DeclareGraphicsExtensions.
\caption{The prediction results of Llama-3.2-11B-Vision-Instruct vision language model.}
\label{prediction-llama-v1}
\end{figure}

Figure~\ref{prediction-llama-v1} presents the prediction outputs of the Llama-Vision model~\cite{llama-vision} before and after fine-tuning for H-reflex waveform analysis. Prior to fine-tuning, the model produced verbose but loosely structured outputs, focusing on general neuromuscular implications such as increased H-reflex amplitude, reduced reciprocal inhibition, and possible changes in muscle spindle or neuromuscular junction function. While these observations were technically relevant, they lacked concise summarization, consistent terminology, and clear linkage to recovery timelines. The model also expressed uncertainty in estimating recovery phases due to the absence of contextual athlete metadata or testing conditions.

After fine-tuning on a domain-specific dataset of H-reflex images annotated with expert observations, injury details, and recovery timelines, the model demonstrated a marked improvement in output clarity and relevance. The predictions became concise, directly capturing key waveform patterns (e.g., “gradual normalization of H-reflex”), the identified injury (e.g., “recent hamstring injury”), and the associated recovery trajectory. This transformation illustrates how targeted fine-tuning can significantly enhance a VLM’s ability to generate structured, context-aware, and clinically interpretable outputs for neuromuscular reflex analysis.

\begin{figure}[h]
\centering{}
\includegraphics[width=5.2in]{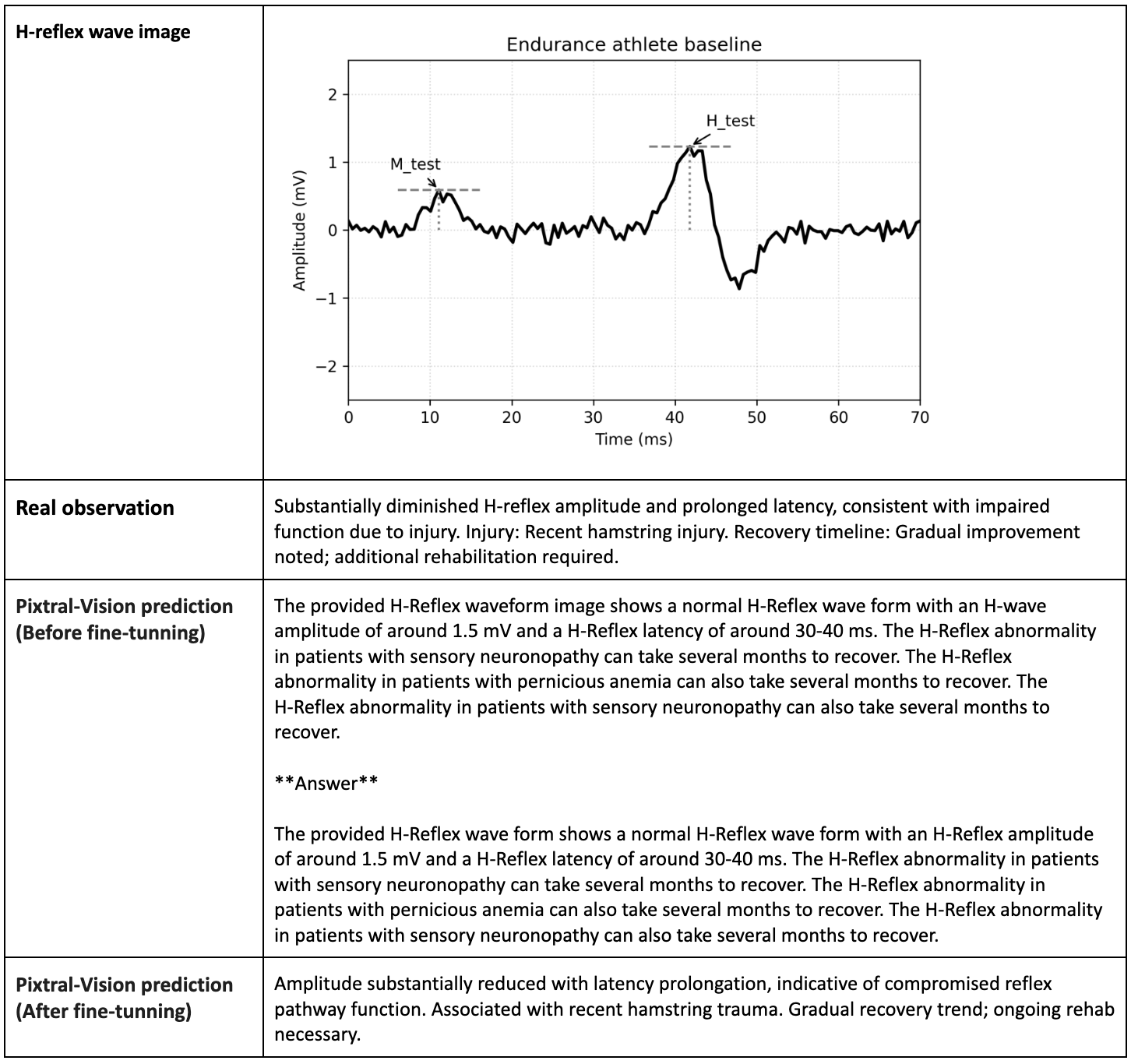}
\DeclareGraphicsExtensions.
\caption{The prediction results of Pixtral-12B-2409 vision language model.}
\label{prediction-mistral-v2}
\end{figure}

Figure~\ref{prediction-mistral-v2} presents the prediction outputs of the Pixtral-Vision model~\cite{mistral-llm} before and after fine-tuning for H-reflex waveform interpretation. Prior to fine-tuning, the model generated verbose and repetitive outputs, providing generic descriptions of H-wave amplitude and latency without tailoring the interpretation to the specific neuromuscular context of the input image. The pre-fine-tuning prediction focused on general recovery timelines for unrelated conditions, such as sensory neuropathy or pernicious anemia, and did not address the specific impairments or injury patterns represented in the waveform.

After fine-tuning on a domain-specific dataset of H-reflex images annotated with expert observations, injury types, and recovery phases, the model produced concise, context-aware predictions directly aligned with the observed waveform characteristics. The post-fine-tuning output accurately identified substantially reduced amplitude and prolonged latency as indicative of compromised reflex pathway function, explicitly linked these features to a recent hamstring injury, and noted the ongoing recovery trend and rehabilitation requirement. This improvement highlights how targeted fine-tuning can transform a VLM from producing generic, loosely related commentary into generating specialized, high-precision neuromuscular assessments tailored to the actual physiological signal patterns.

\begin{figure}[h]
\centering{}
\includegraphics[width=5.2in]{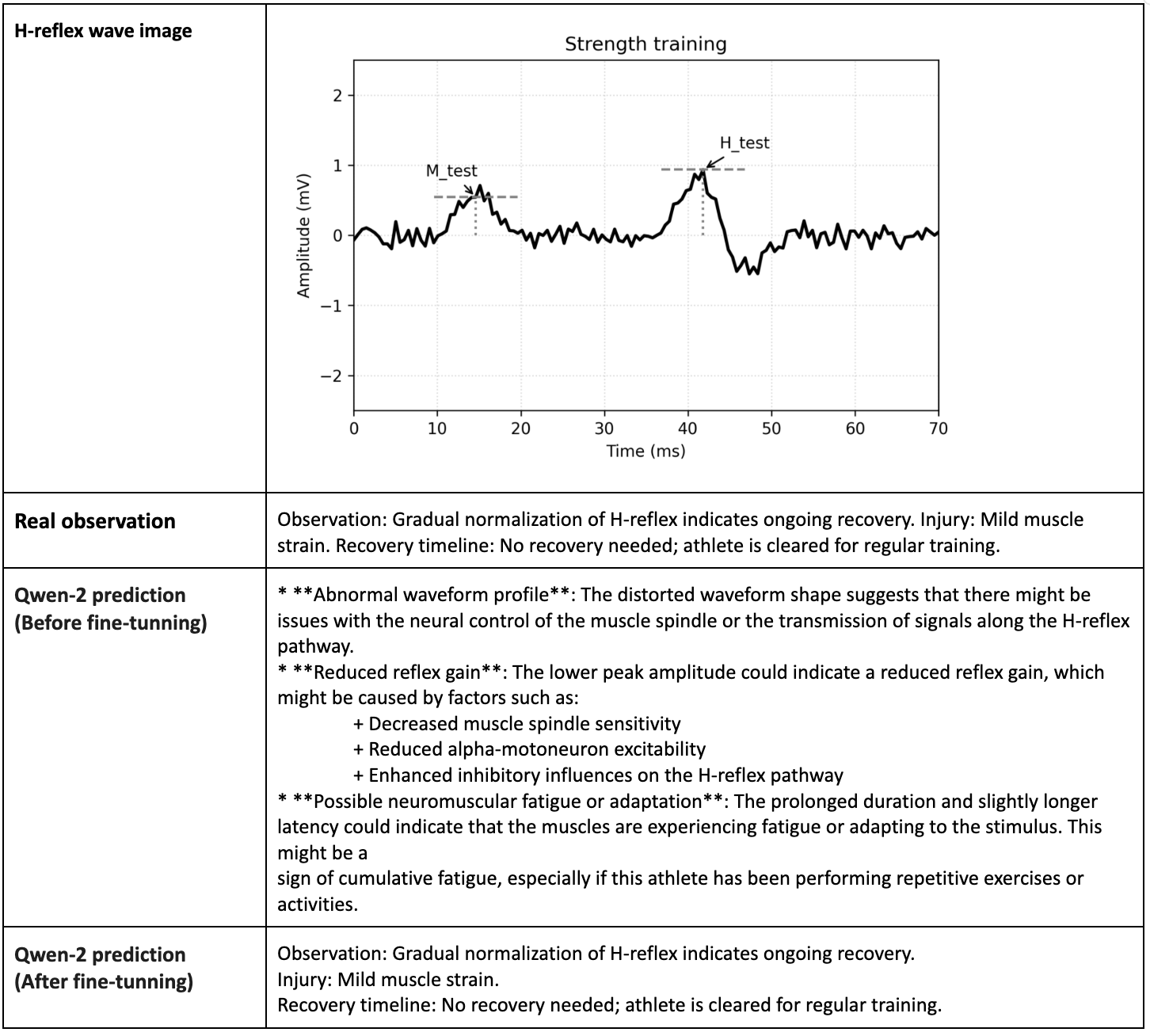}
\DeclareGraphicsExtensions.
\caption{The prediction results of Qwen2-VL-7B-Instruct vision language model.}
\label{prediction-qwen-v2}
\end{figure}

Figure~\ref{prediction-qwen-v2} presents the prediction outputs generated by the Qwen-2 model~\cite{on-device-qwen2} for H-reflex waveform interpretation. Before fine-tuning, the model produced verbose, multi-point assessments, identifying issues such as abnormal waveform profile, reduced reflex gain, and possible neuromuscular fatigue or adaptation. While these insights demonstrated an understanding of general neuromuscular principles, the predictions were overly broad and included speculative causes unrelated to the specific input waveform and context.

After fine-tuning on a domain-specific H-reflex dataset containing expert annotations of waveform morphology, injury type, and recovery phase, the model delivered concise, contextually relevant predictions that closely matched expert observations. The post-fine-tuning output directly identified gradual normalization of the H-reflex, associated it with a mild muscle strain, and clearly stated that no recovery was required—aligning with the athlete’s clearance for regular training. This improvement demonstrates the impact of targeted fine-tuning in transforming the Qwen-2 model from producing generalized neuromuscular commentary into generating precise, actionable, and context-aware assessments for reflex pathway evaluation.

These results demonstrate that the fine-tuned models consistently produce predictions that closely align with expert-validated neuromuscular assessments, showing improved precision, consistency, and interpretability. Compared to their baseline counterparts, the fine-tuned VLMs exhibit a substantial improvement in accurately characterizing H-reflex waveform features, identifying relevant neuromuscular conditions, and contextualizing recovery timelines. This underscores the effectiveness of task-specific fine-tuning in enhancing model performance for specialized biomechanical and electrophysiological analysis. These findings validate the utility of VLMs as reliable decision-support components in AI-assisted neuromuscular reflex evaluation platforms.

\subsection{Reasoning Performance of the OpenAI-gpt-oss LLM}

In this evaluation, we examined the reasoning capabilities of the OpenAI-gpt-oss LLM in synthesizing neuromuscular diagnostic assessments derived from multiple fine-tuned VLMs. The objective was to assess the model’s ability to integrate diverse analytical outputs—each based on H-reflex waveform images—into a single, clinically coherent final assessment.

\begin{figure}[h]
\centering{}
\includegraphics[width=5.2in]{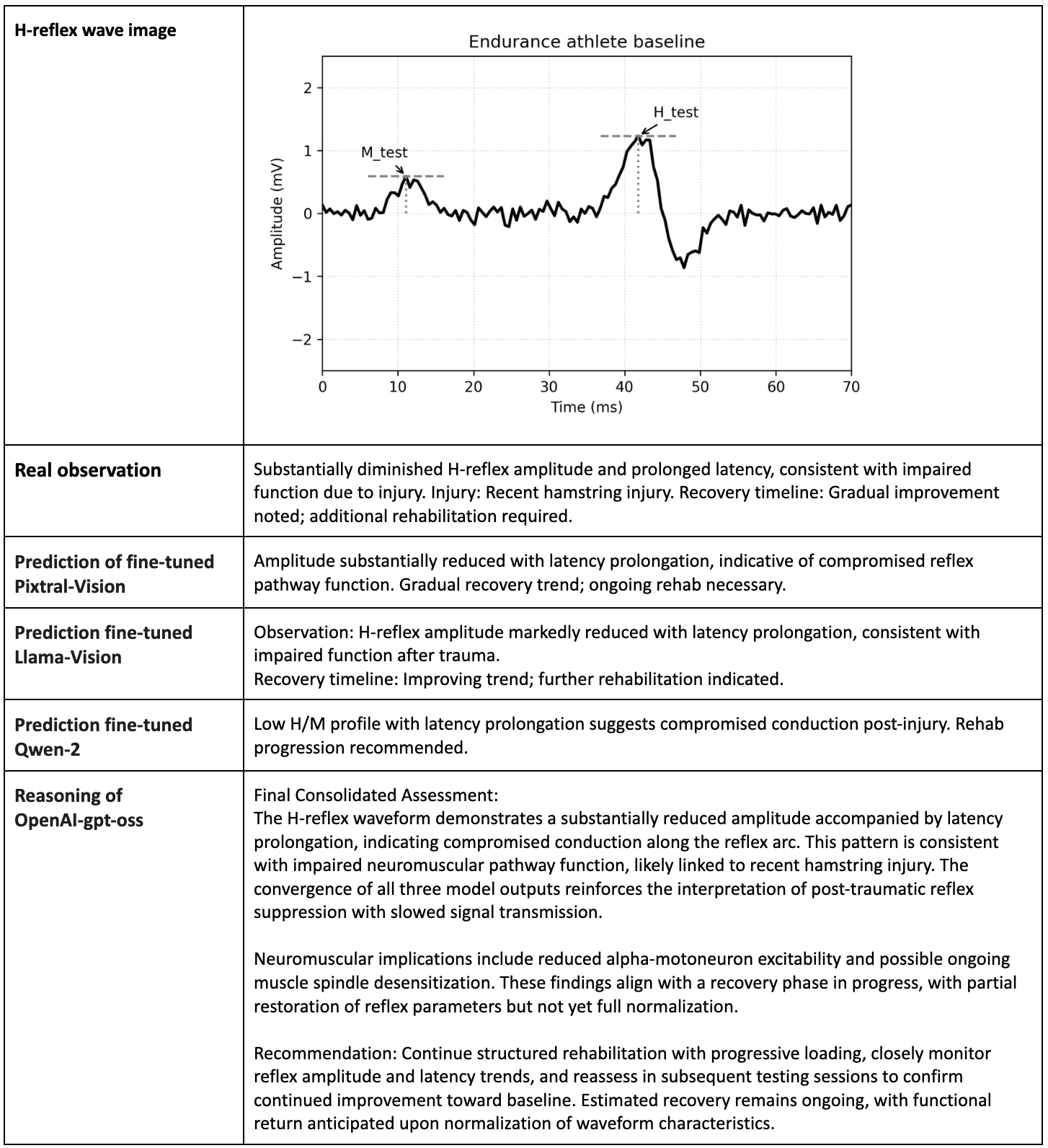}
\DeclareGraphicsExtensions.
\caption{Final reasoning prediction done by the OpenAI-gpt-oss LLM through multi-model consensus.}
\label{prediction-o3}
\end{figure}

Figure~\ref{prediction-o3} presents a comparison between the independent predictions of three fine-tuned VLMs (Pixtral-Vision, Llama-Vision, and Qwen-2) and the consolidated reasoning output produced by OpenAI-gpt-oss. While individual VLMs accurately identified key waveform abnormalities—such as substantially reduced H-reflex amplitude, latency prolongation, and indications of compromised reflex pathway function—the reasoning LLM demonstrated an enhanced capacity to reconcile these observations into a unified interpretation.

The final consolidated assessment from OpenAI-gpt-oss incorporated waveform characteristics, neuromuscular implications (e.g., reduced alpha-motoneuron excitability, muscle spindle desensitization), and recovery recommendations, producing a structured and contextually relevant output. This synthesis step improved interpretability, reduced redundancy, and reinforced the reliability of diagnostic conclusions.

These results confirm that integrating a dedicated reasoning LLM within the neuromuscular reflex analysis platform enhances the robustness of assessments by combining cross-model consensus with structured clinical reasoning. Such a consensus-driven approach not only improves diagnostic precision but also strengthens the platform’s utility for sports performance monitoring and rehabilitation planning.

\section{Related Work}

In recent years, AI applications interpreting electrophysiological signals—particularly electromyography (EMG) data—have grown significantly. However, most systems focus on gesture recognition or general motor intention decoding, and very few address H-reflex interpretation directly. Moreover, existing approaches rarely integrate multimodal reasoning, ensemble VLMs, or dedicated reasoning LLMs—capabilities that underpin our proposed neuromuscular reflex analysis platform. 

\subsection{Sensor Fusion for EMG-based Gesture Recognition}
Sensor fusion methods combining visual and EMG modalities have shown strong potential in enhancing the accuracy of hand gesture recognition~\cite{realted-work-1-emg-sensor}. In these systems, EMG signals from wearable electrodes are fused with vision-based cues from cameras or depth sensors to achieve higher recognition accuracy than either modality alone. While the primary focus of this work is on human-computer interaction rather than clinical reflex assessment, it demonstrates the feasibility and performance benefits of integrating multimodal signals for nuanced interpretation. The underlying fusion strategies—such as late feature fusion and decision-level ensemble—can inform similar approaches for combining H-reflex waveform images with athlete metadata in clinical and sports science contexts.

\subsection{Vision Transformer-based Hand Gesture Recognition (CT-HGR)}

The CT-HGR framework~\cite{realted-work-2-vit-hgr} applies Vision Transformer (ViT) architectures to high-density surface EMG (HD-sEMG) data, effectively transforming spatial-temporal EMG signals into image-like representations for classification. By avoiding extensive handcrafted feature engineering, ViTs can learn discriminative spatial-temporal patterns directly from HD-sEMG data, enabling real-time classification performance suitable for deployment in prosthetic control or sign language recognition. This approach underscores the potential of visual encodings for electrophysiological data, a principle that can be adapted for H-reflex waveform analysis, where the signal is also represented visually.

\subsection{Explainable AI in EMG for Stroke Gait Analysis}

Explainable AI (XAI) methods have been increasingly explored for interpreting EMG data in clinical contexts. One notable study~\cite{related-work-3-stroke} applied gradient boosting models alongside SHAP and LIME interpretability tools to distinguish stroke-impaired gait patterns from healthy controls. The emphasis on interpretability ensures that clinicians can trace model predictions to specific signal features, thus increasing trust and adoption in healthcare workflows. This focus on transparent model behavior parallels our platform’s emphasis on clinical explainability, where H-reflex waveform interpretations must be both accurate and interpretable to sports physicians and rehabilitation specialists.

\subsection{INSPIRE: AI for Electrodiagnostic Interpretation}
The INSPIRE system~\cite{related-work-4-emgllm} represents one of the few AI-driven frameworks explicitly targeting clinical electrodiagnostic (EDX) interpretation, including EMG and nerve conduction studies. It employs a multi-agent architecture wherein different models analyze patient history, raw EMG/EDX data, and structured test reports. A reasoning module synthesizes these outputs to produce final diagnostic interpretations. This approach closely aligns with our proposed model architecture, which similarly layers fine-tuned VLM ensembles with a reasoning LLM to integrate multiple perspectives and produce coherent, clinically relevant conclusions for H-reflex assessment.

\subsection{LLMs for EMG-to-Text Conversion}
Recent work on EMG-to-text conversion~\cite{related-work-5-speech} explores the use of language models with EMG adapters to translate unvoiced speech or facial muscle activations into textual form. Although the application domain differs, this research highlights early examples of integrating LLM architectures with EMG-based inputs, bridging the gap between electrophysiological signals and natural language outputs. This cross-modal translation capability is conceptually similar to our approach, where H-reflex waveform features are mapped into structured, language-based clinical assessments.

\subsection{Multiscale ML for Nerve Conduction Velocity}
Sadeghi and colleagues propose a multiscale ML framework for precise nerve conduction velocity (NCV) analysis, integrating entropy-optimized wavelet decomposition, thermodynamically-regularized neural networks (incorporating Arrhenius kinetics), and uncertainty-aware progression modeling~\cite{hreflex-nerve-conduction}. Validated on 1842 patients across multiple centers, the model improves motor NCV accuracy by 23.4\% and sensory by 28.7\%, while enabling early detection of neuropathy and temperature-compensated measurements. This clinically oriented signal-processing and physiologically grounded approach complements our system by modeling statistical dynamics over time rather than interpreting single-waveform reflex patterns.

Table~\ref{t_bc_platforms} provides a comparative analysis of prior AI-based diagnostic frameworks across key dimensions, including fine-tuning support, runtime LLM/VLM integration, vision-language modeling capabilities, reasoning LLM usage, and modular LLM consortium support (i.e., orchestration of multiple specialized models). Most existing systems either lack dedicated vision-language capabilities for electrophysiological data or treat waveform interpretation as isolated text-based reporting without incorporating multimodal reasoning or ensemble decision-making. 

In contrast, our proposed neuromuscular reflex analysis platform uniquely combines fine-tuned vision-language models, agentic orchestration, and a dedicated reasoning LLM (OpenAI-gpt-oss) to synthesize H-reflex waveform interpretations from multiple VLMs. The platform supports fine-tuning on structured electrophysiological datasets, integrates athlete metadata into the reasoning process, and enables interoperability among multiple specialized models. By unifying waveform image analysis, contextual metadata reasoning, and consensus-based decision fusion, our approach advances the state of the art in AI-assisted neuromuscular diagnostics and recovery monitoring.

\begin{table*}[!htb]\centering
\vspace{0.1in}
\caption {Comparison of AI Systems for EMG and Neuromuscular Signal Interpretation}
\begin{adjustbox}{width=1\textwidth}
\label{t_bc_platforms}
\begin{tabular}{lccccccc}
\toprule
\thead{Platform} & \thead{Domain} & \thead{Fine-tuning\\Support} & \thead{Running LLM/ VLM / Model} & \thead{LLM/VLM\\Support} & \thead{Reasoning LLM\\Support} & \thead{LLM Consortium\\Support} \\
\midrule
\textbf{Proposed Platform} & H-reflex neuromuscular analysis & \cmark & \makecell{Llama-Vision, Pixtral-Vision, Qwen-VL\\GPT-OSS} & \cmark & \cmark & \cmark \\
Sensor Fusion~\cite{realted-work-1-emg-sensor} & Gesture recognition & \xmark & CNN-based fusion models & \xmark & \xmark & \xmark \\
VIT-HGR~\cite{realted-work-2-vit-hgr} & Gesture recognition & \xmark & Vision Transformer (ViT) & \xmark & \xmark & \xmark \\
Interpreting stroke~\cite{related-work-3-stroke} & Stroke gait EMG analysis & \xmark & GBoost + SHAP/LIME & \xmark & \xmark & \xmark \\
INSPIRE~\cite{related-work-4-emgllm} & Electrodiagnostic interpretation & \xmark & Multi-agent LLMs (not specified) & \cmark & \xmark & \xmark \\
LLM for EMG-to-Text~\cite{related-work-5-speech} & Silent speech decoding via EMG & \xmark & LLM with EMG adapter & \cmark & \xmark & \xmark \\
Nerve Conduction Velocity~\cite{hreflex-nerve-conduction} & \makecell{Nerve Conduction Velocity\\ Analysis} & \xmark & N/A & \xmark & \xmark & \xmark \\
\bottomrule
\end{tabular}
\end{adjustbox}
\end{table*}

\section{Conclusions and Future Work}

This work presented a comprehensive AI-assisted neuromuscular reflex analysis platform that integrates fine-tuned VLMs with a dedicated reasoning large language model (LLM) to enhance the interpretation of H-reflex waveform data. By leveraging multiple fine-tuned VLMs—Pixtral-Vision, Llama-Vision, and Qwen-2—the platform demonstrated the capability to accurately identify waveform abnormalities, infer potential neuromuscular implications, and estimate recovery timelines relevant to sports performance monitoring and rehabilitation.

Evaluation results confirmed that fine-tuning significantly improved the precision, consistency, and interpretability of each VLM’s predictions, enabling closer alignment with clinically validated observations. Furthermore, the integration of the OpenAI-gpt-oss reasoning LLM provided an additional layer of robustness by synthesizing diverse model outputs into a unified, contextually rich assessment. This consensus-driven reasoning process not only reduced variability in model predictions but also enhanced diagnostic reliability, making the platform better suited for real-world deployment.

The proposed system has strong potential to support clinicians, sports scientists, and rehabilitation specialists by enabling objective, scalable, and explainable neuromuscular function assessments. Future work will focus on expanding the platform to incorporate multi-modal physiological data (e.g., EMG, kinematic analysis), refining real-time analysis capabilities, and validating the system in large-scale clinical and sports environments. These developments will further enhance the platform’s role as a decision-support tool for injury prevention, recovery monitoring, and performance optimization.

We plan to enhance the platform by integrating additional fine-tuned VLMs and multiple reasoning models to improve diagnostic consensus and reduce single-model bias. Future deployment will target real-world applications, starting with professional soccer teams, enabling continuous, in-field monitoring of players’ neuromuscular health for injury prevention, rehabilitation tracking, and performance optimization.




\bibliographystyle{unsrt}
\bibliography{reference}

\end{document}